\documentclass{article} 
\usepackage{iclr2026_conference,times}

\usepackage{amsmath,amsfonts,bm}

\def\eqref#1{equation~\ref{#1}}

\def\1{\bm{1}}

\DeclareMathAlphabet{\mathsfit}{\encodingdefault}{\sfdefault}{m}{sl}
\SetMathAlphabet{\mathsfit}{bold}{\encodingdefault}{\sfdefault}{bx}{n}

\usepackage{hyperref}
\usepackage{url}

\usepackage{algorithm}
\usepackage{algorithmic}

\usepackage[utf8]{inputenc} 
\usepackage[T1]{fontenc}    
\usepackage{url}            
\usepackage{booktabs}       
\usepackage{amsfonts}       
\usepackage{nicefrac}       
\usepackage{microtype}      
\usepackage{xcolor}         

\usepackage[utf8]{inputenc} 
\usepackage[T1]{fontenc}    
\usepackage{url}            
\usepackage{booktabs}       
\usepackage{amsfonts}       
\usepackage{nicefrac}       
\usepackage{microtype}      
\usepackage{xcolor}         
\usepackage{amsmath}
\usepackage{amsthm}
\usepackage{mathtools}
\usepackage{subcaption}
\usepackage[inkscapearea=page]{svg}
\usepackage{tabularx}
\usepackage{enumitem}

\newtheorem{theorem}{Theorem}

\newtheorem{definition}{Definition}

\title{Automating the Refinement of Reinforcement Learning Specifications}

\author{
Tanmay Ambadkar\thanks{Department of EECS, The Pennsylvania State University} \\
The Pennsylvania State University \\
\texttt{tsa5252@psu.edu} \\
\And
Đorđe Žikelić\thanks{School of Computing and Information Systems, Singapore Management University} \\
Singapore Management University \\
\texttt{dzikelic@smu.edu.sg} \\
\And
Abhinav Verma\thanks{Department of EECS, The Pennsylvania State University} \\
The Pennsylvania State University \\
\texttt{verma@psu.edu}
}

\newcommand{\toolname}{\textsc{AutoSpec}}

\iclrfinalcopy 
\begin{document}

\maketitle
\begin{abstract}
Logical specifications have been shown to help reinforcement learning algorithms in achieving complex tasks. However, when a task is under-specified, agents might fail to learn useful policies. In this work, we explore the possibility of improving coarse-grained logical specifications via an exploration-guided strategy. We propose \toolname{}, a framework that searches for a logical specification refinement whose satisfaction implies satisfaction of the original specification, but which provides additional guidance therefore making it easier for reinforcement learning algorithms to learn useful policies. \toolname{} is applicable to reinforcement learning tasks specified via the SpectRL specification logic. We exploit the compositional nature of specifications written in SpectRL, and design four refinement procedures that modify the abstract graph of the specification by either refining its existing edge specifications or by introducing new edge specifications. We prove that all four procedures maintain specification soundness, i.e. any trajectory satisfying the refined specification also satisfies the original. We then show how \toolname{} can be integrated with existing reinforcement learning algorithms for learning policies from logical specifications. Our experiments demonstrate that \toolname{} yields promising improvements in terms of the complexity of control tasks that can be solved, when refined logical specifications produced by \toolname{} are utilized.
\end{abstract}

\section{Introduction}
\label{sec:intro}

Reinforcement Learning (RL) algorithms have made tremendous strides in recent years~\cite{sutton2018reinforcement, SilverHMGSDSAPL16, MnihKSRVBGRFOPB15, LevineFDA16}. However, most algorithms assume access to a scalar reward function that must be carefully engineered to make environments amenable to RL—a practice known as reward engineering~\cite{rewardengineering}. This creates challenges in applying RL to new environments where useful reward functions are hard to construct. Furthermore, scalar Markovian rewards cannot provide sufficient feedback for certain tasks~\cite{abel2021reward, bowling2023settling}, leading to growing interest in non-Markovian reward functions~\cite{DBLP:journals/corr/LiVB16, dirl, alur2023specification}.

To make non-Markovian rewards tractable, it is standard to represent them via logical specification formulas that capture the intended task. These approaches, known as specification-guided reinforcement learning~\cite{AksarayJKSB16,LiVB17,IcarteKVM18,spectrl,dirl}, derive reward functions from logical specifications. However, this creates two challenges: (i) providing specifications granular enough to guide RL algorithms, and (ii) defining accurate labeling functions mapping environment states to specification predicates. Users often create coarse specifications or labeling functions that, while logically correct, provide insufficient guidance for learning.

\begin{figure*}[!htb]
    \centering
    \includegraphics[width = 0.8\textwidth]{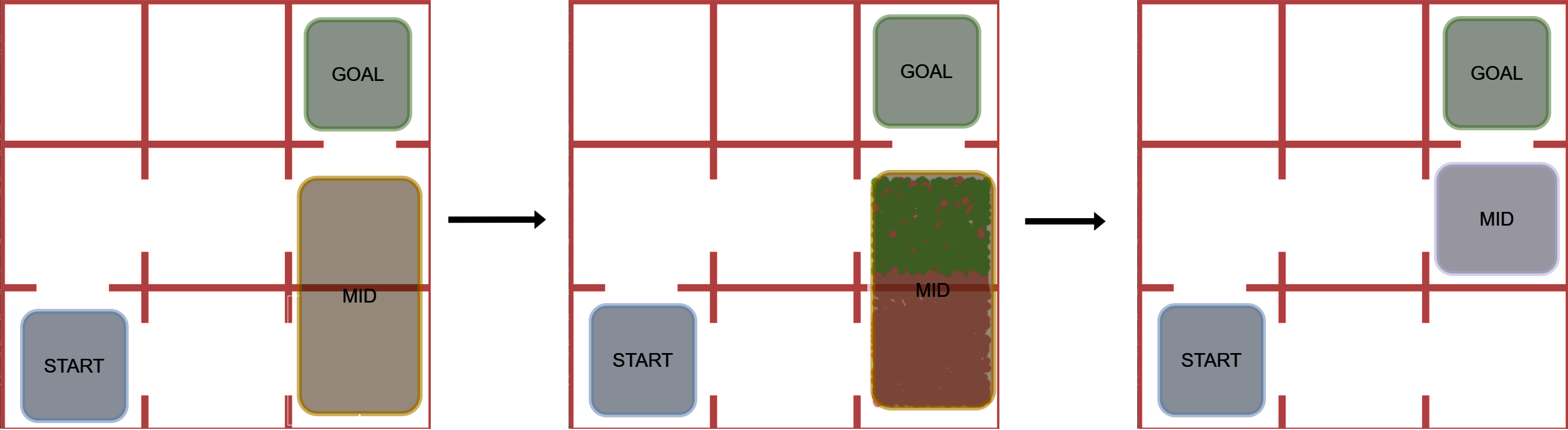}
    \caption{Example of refinement by \toolname{} in a 9-rooms environment. The original MID-node region includes a trap state from which recovery is impossible. The refined specification excludes this trap, enabling the agent to learn a policy with higher satisfaction probability.}
    \label{fig:example}
\end{figure*}

We present \toolname{}, a framework for automatically refining coarse specifications without user intervention. 
We say that a logical specification is \textit{coarse} (or \textit{under-specified}) if its predicate labelings or logical structure are too coarse to allow specification-guided RL algorithms to translate logical specifications into reward functions that allow for effective learning of RL policies. 
\toolname{} starts with an initial logical specification, translates it to a reward function, and attempts to learn a policy. If the learned policy's performance is unsatisfactory, \toolname{} identifies which specification components cause learning failures and automatically refines both the specification formula and labeling function. The refined specification's satisfaction implies the original's satisfaction while providing additional structure for learning. This process repeats until a satisfactory policy is learned.

\toolname{} works with SpectRL specifications; boolean and sequential combinations of reach-avoid tasks~\cite{spectrl}. Any SpectRL specification decomposes into an abstract graph where edges specify reach-avoid tasks~\cite{dirl}. \toolname{} identifies problematic edges and applies targeted refinements: either modifying the labeling function for regions (Figure~\ref{fig:example}) or restructuring the graph to add alternative paths. 
These problematic edges are identified by employing an \textit{exploration-guided strategy} that utilizes empirical trajectory data to identify edges in the abstract graph whose reach-avoid tasks (i.e.~initial, target or unsafe regions) make it hard to learn a good RL policy. For instance, in Figure ~\ref{fig:example}, the initial \texttt{MID} region of the \texttt{MID}-\texttt{GOAL} edge in the abstract graph is under-specified, as it overlaps with a trap state. The trap state is not immediately obvious as there is a path from \texttt{MID} to \texttt{GOAL}. By analyzing explored traces, \textsc{AutoSpec} identifies problematic start states in \texttt{MID} and refines the region to exclude the trap (as shown in Figure~\ref{fig:example}), thereby refining the logical reach-avoid specification associated to the \texttt{MID}-\texttt{GOAL} edge.

We prove that all refinements maintain soundness, where satisfaction of the refined specification implies satisfaction of the original. \toolname{} integrates with existing SpectRL-compatible algorithms as demonstrated with \textsc{DiRL}~\cite{dirl} and \textsc{LSTS}~\cite{lsts}.


Our contributions:
\begin{enumerate}
    \item A framework for automated refinement of logical RL specifications with four refinement procedures, all with formal soundness guarantees (Section~\ref{sec:refinement}).
    \item Integration with existing specification-guided RL algorithms, enabling them to solve tasks with coarse specifications (Section~\ref{sec:refinement}).
    \item Empirical demonstration that \toolname{} enables learning from specifications that existing methods cannot handle (Section~\ref{sec:experiments_main}).
\end{enumerate}

\smallskip\noindent{\bf Related work.} Recent years have seen substantial progress in solving RL tasks specified via logical specifications \cite{AksarayJKSB16,LiVB17,IcarteKVM18,CamachoIKVM19,GiacomoIFP19,HasanbeigKA22,HasanbeigKAKPL19,HahnPSSTW19,spectrl,dirl,0005T19}. Many works consider different fragments of Linear Temporal Logic (LTL) or their variants for specifying RL tasks. \cite{IcarteKVM18,CamachoIKVM19} consider tasks that can be specified using deterministic finite automata (DFA) and solve them by {\em reward machines}, which decompose these tasks and translate them into a reward function. The reward function can then be used to train existing RL algorithms. \cite{LiVB17} considers a variant of LTL called TLTL for specifying tasks and propose a method for translating these specifications into continuous reward functions. \cite{HasanbeigKA22,HasanbeigKAKPL19, HahnPSSTW19} study the translation of tasks specified in LTL into reward functions. \cite{alur2022framework} examines the theoretical questions related to the translation of logical specifications into reward functions.

\cite{spectrl} defines the specification language SpectRL, a finitary fragment of LTL and provides justification for using this language to define specifications for RL tasks. A compositional method that decomposes SpectRL specifications into an abstract graph and constructs a reward function for each abstract graph edge was proposed in \cite{dirl}. The approach by \cite{perfectrm} focuses on discovering optimal reward structures through environmental exploration and reward analysis. Compositional methods are further explored by \cite{subtask-removal}, who propose removing unfulfillable subtasks, and \cite{neary2023verifiable}, who introduce verification techniques to certify learned policies. \cite{vzikelic2024compositional} propose \textsc{Claps}, a compositional method for learning neural network policies with formal guarantees on the satisfaction of SpectRL specifications, thus advancing the applicability to safety-critical RL applications by utilizing prior methods for learning reach-avoid policies with formal guarantees~\cite{LechnerZCH22,ZikelicLHC23,ChatterjeeHLZ23}.

Recent advancements also include LTL2Action \cite{ltl2action}, which translates LTL specifications into sequences of tasks for RL agents. Other recent approaches, such as ~\cite{qiu2023instructing} and DeepLTL~\cite{jackermeier2025deepltl}, leverage goal-conditioned RL and automata-based architectures to solve complex LTL and $\omega$-regular tasks zero-shot or in multi-task settings. Trainify~\cite{trainify} employs counterexample-guided abstraction and refinement (CEGAR) to iteratively improve policies by addressing failure cases identified through counterexamples. However, these works primarily focus on learning policies for fixed, well-defined specifications. In contrast, \textsc{AutoSpec} studies the problem of automated logical specification refinement towards improving reward functions obtained by translation from coarse logical RL specifications. Thus, our work is complementary to the works on RL from logical specifications and can be integrated into off-the-shelf specification-guided RL algorithms to improve the performance of learned agents.

\section{Preliminaries}
\label{sec:prelims}


{\bf MDPs.} A Markov Decision Process (MDP) is a tuple $M = (S, A, P, R, \gamma)$, where $S \subseteq \mathbb{R}^n$ is the state space, $A \subseteq \mathbb{R}^m$ is the action space, $P: S \times A \times S \rightarrow [0, 1]$ is the probabilistic transition function, $R$ is the (possibly non-Markovian) reward function, and $\gamma$ is the discount factor. Let $\eta: S \rightarrow [0, 1]$ be the initial state distribution.  A trajectory $\zeta$ in $M$ is a sequence of states and actions $\zeta = s_0, a_0, s_1, a_1, \ldots$ where $s_i \in S$ and $a_i \in A$. We use $\mathcal{Z}$ to denote the set of all trajectories in $M$ and $\mathcal{Z}_f$ to denote the set of all finite trajectories in $M$, which are finite prefixes of trajectories ending in states. A (pure) policy $\pi: \mathcal{Z}_f \rightarrow A$ assigns an action to each finite trajectory, and a non-Markovian reward $R: \mathcal{Z}_f \rightarrow \mathbb{R}$ assigns a reward to a finite trajectory. The MDP $M$ under any policy $\pi$ gives rise to a probability space over the set of all trajectories in the MDP~\cite{Puterman94}. We use $\mathbb{P}^\pi$ and $\mathbb{E}^\pi$ to denote the probability measure and the expectation operator in this probability space, respectively. 

\noindent{\bf Logical specifications for Reinforcement Learning.} In this work, we are solving RL tasks defined by logical specifications. Formally, a {\em logical specification} (or, simply, a {\em specification}) is a boolean function $\phi: \mathcal{Z} \rightarrow \{\textrm{true},\textrm{false}\}$ which specifies whether a trajectory in the MDP satisfies the specification. We write $\zeta\models\phi$ whenever a trajectory $\zeta$ satisfies the specification $\phi$. The objective of a specification-guided RL task is to find a policy $\pi^*$ that maximizes the probability of satisfying the given specification $\phi$, i.e. $\pi^* \in \textit{argmax}_\pi \mathbb{P}^\pi [\zeta\models\phi]$. Specification-guided RL algorithms use the specification to create a dense reward that guides the policy search, and therefore outperform algorithms that cannot leverage the specification for learning and instead require manual reward engineering \cite{dirl, lsts}.

\noindent{\bf SpectRL specification logic.} We consider RL tasks specified in the SpectRL specification logic. SpectRL~\cite{spectrl} is a fragment of Linear Temporal Logic (LTL) which consists of all boolean and sequential combinations of reach-avoid tasks. Formally, a specification in SpectRL is defined in terms of {\em predicates} and {\em specification formulas}. An atomic predicate is a function $a: S \rightarrow \{\textrm{true},\textrm{false}\}$ which defines a set of states that satisfy the atomic predicate. A predicate is a boolean combination of atomic predicates, i.e.~$b := a \,\mid\, b_1\land b_2 \,\mid\, b_1\lor b_2$, where $a$ is an atomic predicate and $b_1$ and $b_2$ are predicates. Specification formulas in SpectRL are defined by the grammar
\begin{equation}\label{eq:grammar}
 \phi := \textrm{achieve } b \,\mid\, \phi_1 \textrm{ ensuring } b \mid \phi_1 ; \phi_2 \mid \phi_1 \textrm{ or } \phi_2
\end{equation}
where $b$ is a predicate and $\phi_1$ and $\phi_2$ are specification formulas. Intuitively, ''$\textrm{achieve } b$'' requires the agent to reach a state in which the predicate $b$ is satisfied. The clause ''$\phi_1 \textrm{ ensuring } b$'' requires the agent to satisfy the specification $\phi$ while only visiting states in which the predicate $b$ is satisfied. The clause ''$\phi_1 ; \phi_2$'' requires the agent to first satisfy specification $\phi_1$ and then satisfy specification $\phi_2$. The clause ''$\phi_1 \textrm{ or } \phi_2$'' requires satisfaction of at least one of $\phi_1$ or $\phi_2$. See~\cite{spectrl} for the formal definition of the semantics of each clause.

\noindent{\bf Abstract graphs for SpectRL specifications.} It was shown in~\cite{dirl} that each specification written in the SpectRL specification logic can be translated into an equivalent abstract graph. An {\em abstract graph} is a directed acyclic graph (DAG) whose vertices represent sets of MDP states and whose edges are annotated with sets of safe MDP states. Hence, each abstract graph edge defines a {\em reach-avoid specification}, where the task is to reach the set of states defined by the target vertex of the edge starting from the set of states defined by the source vertex of the edge, while staying within the set of safe states defined by the edge.

\begin{definition}[Abstract graph]\label{def:abstractgraph}
An {\em abstract graph} $G = (V, E, \beta, s, t)$ is a DAG, where $V$ is a finite set of vertices, $E$ is a finite set of edges, $\beta: V\cup E\rightarrow \mathcal{B}(S)$ is a labeling function that maps each vertex and each edge to a subset of the MDP states $S$, $s\in V$ is the source vertex and $t\in V$ is the target vertex. Furthermore, we require that $\beta(s) = \text{support}(\eta)$ is the support of the initial state distribution $\eta$ of the MDP.
\end{definition}

Given a trajectory $\zeta$ in the MDP and an abstract graph $G = (V, E, \beta, s, t)$, we say that $\zeta$ satisfies {\em abstract reachability} for $G$ (written $\zeta\models G$) if it gives rise to a path in $G$ that traverses $G$ from $s$ to $t$ and satisfies the reach-avoid specifications of every traversed edge. It was shown in~\cite{dirl} that, given any SpectRL specification $\phi$, one can construct an abstract graph $G$ such that $\zeta\models\phi$ if and only if $\zeta\models G$ holds for each trajectory $\zeta$ in the MDP. Hence, solving an RL task for a SpectRL specification reduces to solving an abstract reachability task in the abstract graph $G$.

\noindent{\bf Problem statement.} Given an MDP $M$ and a SpectRL specification $\phi$, our goal is to learn a policy $\pi$ such that the probability $\mathbb{P}^\pi[\zeta \models \phi]$ of a trajectory satisfying the specification is maximized.

\noindent{\bf Specification refinement.} In order to solve this problem, we will utilize a common approach in specification-guided RL, to first translate the logical specification $\phi$ to a (non-sparse) reward function and then learn a policy by using existing RL algorithms with this reward function. However, if the probability of the specification being satisfied under the learned policy is unsatisfactory (i.e.~below some desired probability threshold $p \in [0,1]$), we will then refine the logical specification $\phi$ into a new SpectRL specification $\phi_r$. We will then repeat the above process until the probability of the specification being satisfied under the learned policy becomes satisfactory.

\begin{definition}[Specification refinement]\label{def:refinement}
Given two logical specifications $\phi$ and $\phi_r$, we say that $\phi_r$ {\em refines} $\phi$, if any MDP trajectory that satisfies the refined specification $\phi_r$ also satisfies the specification $\phi$. That is, if for an MDP trajectory $\zeta$ we have $(\zeta \models \phi_r) \implies (\zeta \models \phi)$.
\end{definition}

\section{Automated Refinement of RL Specifications}
\label{sec:refinement}

We now present \toolname{}, a framework for automated refinement of logical specifications in RL tasks. The key insight is that specification failures often stem from identifiable issues that can be systematically addressed: overly broad target regions, insufficient safety constraints, missing waypoints, or lack of alternative paths. When a specification-guided RL algorithm $\mathcal{A}$ fails to learn a satisfactory policy for specification $\phi$, \toolname{} identifies which components caused the failure and applies targeted refinements to improve learnability while maintaining soundness.

\toolname{} operates as a wrapper around any SpectRL-compatible algorithm. It monitors the learning process, and when a policy $\pi$ fails to satisfy the specification with probability at least $p \in [0,1]$ (a user-provided threshold), it computes a refined specification $\phi_r$ such that: (1) satisfaction of $\phi_r$ implies satisfaction of $\phi$ (soundness), and (2) $\phi_r$ provides additional structure that makes it easier to learn. This refined specification is returned to algorithm $\mathcal{A}$ to continue learning. Through this iterative refinement process, \toolname{} enables solving RL tasks with coarse specifications that would otherwise be unlearnable.

\noindent{\bf Overview of \toolname{}.} Algorithm~\ref{alg:autospecpaths} shows the complete \toolname{} framework. The algorithm takes as input an MDP $M$, a SpectRL specification $\phi$, a satisfaction threshold $p$, and any specification-guided RL algorithm $\mathcal{A}$. It first translates $\phi$ into an abstract graph $G$ and uses $\mathcal{A}$ to learn policies for the graph edges. For each edge $e$ where $\mathcal{A}$ learned a policy but that policy fails to achieve satisfaction probability $p$, \toolname{} applies four refinement procedures in sequence: SeqRefine, AddRefine, PastRefine, and OrRefine. This ordering reflects increasing levels of structural modification; from local predicate adjustments to graph topology changes. The first refinement that successfully improves performance above threshold $p$ is applied, the graph is updated, and policies are relearned before proceeding to the next edge.

\begin{algorithm}[!htb]
\caption{\toolname{} }\label{alg:autospecpaths}
\begin{algorithmic}
\REQUIRE MDP $M$, specification $\phi$, threshold $p \in [0,1]$, spec-guided RL algorithm $\mathcal{A}$
\STATE $G \gets$ abstract graph corresponding to $\phi$
\STATE $\Pi \gets \mathcal{A}(G)$ [set of policies for edges in $G$ learned by algorithm $\mathcal{A}$] 
\FOR{$e = u \rightarrow u'$ an edge in $G$}
\STATE $\pi_e \in \Pi \gets$ policy learned for edge $e$ (Null if $\mathcal{A}$ does not learn a policy for edge $e$)
\IF{$\pi_e$ is not Null and $\mathbb{P}(\pi_e) < p$}
\STATE $\zeta \gets$ sampled trajectories of the system
\IF {\textsc{LearnPolicy}(e,\textsc{SeqRefine}(e,$G$,$\zeta$))$ > p$}
    \STATE $G \gets$ \textsc{SeqRefine}(e,$G$,$\zeta$)
\ELSIF {\textsc{LearnPolicy}(e,\textsc{AddRefine}(e,$G$,$\zeta$))$ > p$}
    \STATE $G \gets$ \textsc{AddRefine}(e,$G$,$\zeta$)
\ELSIF{\textsc{LearnPolicy}(e,\textsc{PastRefine}(e,$G$,$\zeta$))$ > p$}
    \STATE $G \gets$ \textsc{PastRefine}(e,$G$,$\zeta$)
\ELSIF{\textsc{LearnPolicy}(e,\textsc{OrRefine}(e,$G$,$\zeta$))$ > p$}
    \STATE $G \gets$ \textsc{OrRefine}(e,$G$,$\zeta$)
\ENDIF
\STATE $\Pi \gets \mathcal{A}(G)$ [set of policies for updated abstract graph $G$] 
\ENDIF
\ENDFOR

\STATE \textbf{Return} $G$ and $\Pi$

\end{algorithmic}
\end{algorithm}



\toolname{} iterates through all edges $e = u \rightarrow u'$ of the abstract graph $G$ for which the specification-guided RL algorithm $\mathcal{A}$ has learned a policy but for which the reach-avoid task satisfaction probability is below the provided probability threshold $p$. For each such edge, \toolname{} performs four refinement procedures that focus on different possible reasons for the edge $e = u \rightarrow u'$ being challenging for learning a satisfactory policy. 

SeqRefine, which is invoked first, tries to locally refine the problematic edge $e = u \rightarrow u'$ by using predicate refinement techniques to refine the labeling function at the target region associated to the vertex $u'$ and the safety region associated to the edge $e$. If SeqRefine fails to improve performance above threshold, \toolname{} invokes AddRefine which attempts to add a waypoint (i.e.~a new abstract graph vertex) between the vertices $u$ and $u'$, making path-finding easier. If AddRefine also fails, \toolname{} invokes PastRefine which tries to refine the source node $u$. Finally, if other refinement procedures fail, \toolname{} invokes OrRefine which aims to find alternative paths to $u'$. 

After each attempted refinement, an off-the-shelf RL algorithm (\textsc{LearnPolicy} in Algorithm~\ref{alg:autospecpaths}) is used to estimate the satisfaction probability of the refined edge. When a refinement succeeds in achieving satisfaction probability above $p$, the refined abstract graph $G$ is updated and \toolname{} applies the specification-guided RL algorithm $\mathcal{A}$ to learn a new set of edge policies $\Pi$ for the entire graph. At the end, the final abstract graph $G$ corresponds to the refined specification $\phi_r$ of the input specification $\phi$.

\subsection{Specification Refinement Subprocedures}
\label{subsec:refine}

We now define the four specification refinement subprocedures used by \toolname{} in Algorithm~\ref{alg:autospecpaths}. Each procedure addresses a specific type of specification inadequacy, and they are applied in order of increasing structural modification to the abstract graph. The detailed pseudocodes are provided in the Appendix. Once a problematic edge $e = u \rightarrow u'$ is identified (an edge with satisfaction probability below threshold $p$), \toolname{} samples trajectories $\zeta$ using the learned policy, where the number of trajectories is an algorithm hyperparameter.

\textbf{SeqRefine: Refining Predicates.} 
The first refinement subprocedure addresses overly coarse predicates in the reach and avoid conditions. For edge $e = u \rightarrow u'$, SeqRefine refines both the target predicate $b = \beta(u')$ and the safety predicate $c = \beta(e)$ by calling two subprocedures:

\begin{figure}[!t]
    \centering
    
    \begin{subfigure}[b]{0.18\textwidth}
        \centering
        \includegraphics[width=\linewidth]{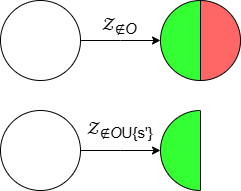}
        \caption{ReachRefine}
        \label{fig:reach_refine}
    \end{subfigure}
    \hfill 
    \begin{subfigure}[b]{0.23\textwidth}
        \centering
        \includegraphics[width=\linewidth]{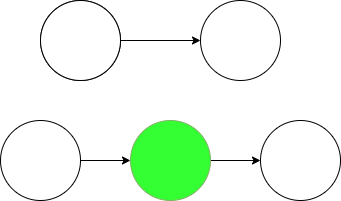}
        \caption{AddRefine}
        \label{fig:add_refine}
    \end{subfigure}
    \hfill 
    \begin{subfigure}[b]{0.23\textwidth}
        \centering
        \includegraphics[width=\linewidth]{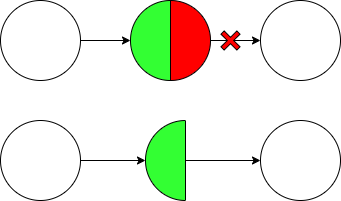}
        \caption{PastRefine}
        \label{fig:past_refine}
    \end{subfigure}
    \hfill 
    \begin{subfigure}[b]{0.23\textwidth}
        \centering
        \includegraphics[width=\linewidth]{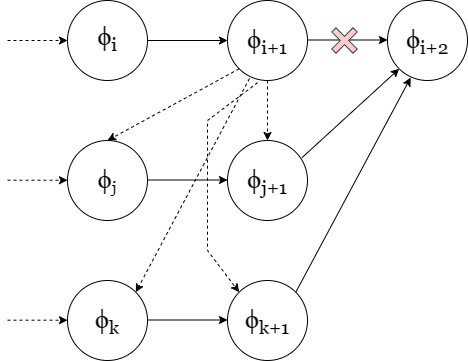}
        \caption{OrRefine}
        \label{fig:or_refinement}
    \end{subfigure}
    
    \caption{Illustrations of abstract graph refinement processes. 
    (a) ReachRefine demonstrating removal of failed part of goal region and addition of unsafe states set to existing avoid. 
    (b) AddRefine demonstrating addition of waypoint between 2 nodes. 
    (c) PastRefine removes part of the source node from where the agent failed (red) to get to the target, and keeps successful start states (green).
    (d) OrRefine shows how alternative paths (dotted lines) to target are constructed using existing specification nodes. 
    }
    \label{fig:graph_refinement}
    \vspace{-10pt}
\end{figure}

\textit{ReachRefine} collects all states along sampled trajectories that successfully reached the goal region $b$. The refined goal region is computed as $b_r = b \cap \text{ConvexHull}(\text{reached states})$, effectively excluding unreachable portions of the original target region. \textit{AvoidRefine} collects states where trajectories entered unsafe regions (complement of $c$). The refined safe region is computed as $c_r = c \setminus \text{ConvexHull}(\text{last } k \text{ unsafe trajectory states})$, where $k$ is a hyperparameter specifying how much of the trajectory tail to remove. This removes demonstrated unsafe areas from the safety region. 

SeqRefine returns a refined graph $G_r$ identical to $G$ except with updated labeling: $\beta_r(u') = b_r$ and $\beta_r(e) = c_r$. This refinement provides more precise guidance by excluding problematic regions discovered through exploration.

\textbf{AddRefine: Introducing Waypoints.}
The second refinement addresses long or complex paths by decomposing them. When direct navigation from $u$ to $u'$ proves difficult, AddRefine introduces an intermediate vertex $u''$ by collecting midpoint states from successful trajectories that reached $\beta(u')$, defining $\beta(u'') = \beta(e) \cap \text{ConvexHull}(\text{midpoints})$, and replacing edge $e = u \rightarrow u'$ with two edges: $e'' = u \rightarrow u''$ and $e' = u'' \rightarrow u'$. This decomposition breaks a challenging long-horizon task into two shorter subtasks that are easier to learn.

\textbf{PastRefine: Partitioning Source Regions.}
The third refinement addresses heterogeneous starting conditions where some initial states in $u$ consistently lead to success while others lead to failure. PastRefine separates trajectories into successful and failing sets based on whether they satisfied edge $e$, then learns a hyperplane separating successful from failing initial states. It creates region $b_r$ containing successful starting states and introduces new vertex $u^*$ with $\beta(u^*) = b_r$ having the same incoming edges as $u$. The refinement replaces problematic edge $e = u \rightarrow u'$ with $e^* = u^* \rightarrow u'$. As shown in Figure~\ref{fig:graph_refinement}(a), this refinement identifies and isolates promising initial conditions while preserving the original vertex $u$ and its connections.

\textbf{OrRefine: Exploiting Alternative Paths.}
The fourth refinement addresses blocked or infeasible direct paths by leveraging the existing graph structure. When the path through edge $e = u \rightarrow u'$ cannot be made satisfactory, OrRefine identifies alternative parents of $u'$ (vertices $u_i$ with existing edges $e_i = u_i \rightarrow u'$), and for each viable $u_i$, adds new edge $e_{new} = u \rightarrow u_i$ with $\beta(e_{new}) = \beta(e)$ and $\beta(e_i) = \beta(e) \cap\beta(e_{i})$. It then tests if the alternative path $u \rightarrow u_i \rightarrow u'$ achieves the threshold. As illustrated in Figure~\ref{fig:graph_refinement}(b), this creates alternative routes to the target using only existing vertices, maintaining all original safety constraints. OrRefine can iteratively explore ancestors of $u_i$ if the direct connection fails.

As shown in Algorithm~\ref{alg:autospecpaths}, any specification-guided RL algorithm that is applicable to SpectRL specifications and that learns policies for edges in the abstract graph can be integrated into the \toolname{} framework. The specification-guided RL algorithm learns policies for edges in the abstract graph, until it is unable to proceed beyond an edge with a sufficient satisfaction probability. We then perform the refinements in \toolname{}, using sampling to estimate the satisfaction probability of each refinement until one is found to exceed the threshold. This refinement is used to create an updated abstract graph and an updated set of edge policies are learned with respect to this graph. 

    
    
    


\subsection{Correctness of \toolname{}}

The following theorem establishes correctness of \toolname{}, showing that the specification $\phi_r$ computed by \toolname{} is indeed a refinement of the input specification $\phi$. The proof, provided in the Appendix, proceeds by proving that each of the four refinement procedures results in a specification refinement.

\begin{theorem}[Correctness of \toolname{}]\label{thm:correctpath}
Given an abstract graph $G$ of a SpectRL specification $\phi$ and an edge $e$, \toolname{} computes a specification $\phi_r$ and returns an abstract graph $G_r$ and an edge $e_r$ such that $\phi_r$ refines $\phi$. That is, for any MDP trajectory $\zeta$, we have $(\zeta \models \phi_r) \implies (\zeta \models \phi)$.
\end{theorem}

\paragraph{Incompleteness of the Specification Refinement Problem.} 
\toolname{} provides \emph{soundness} guarantees -- as shown in Theorem~\ref{thm:correctpath}, the produced specification is \textit{guaranteed} to be a refinement of the original specification as in Definition~\ref{def:refinement}, and every trajectory that satisfies the refined satisfaction must also satisfy the original specification. However, \toolname{} does not provide completeness guarantees. This is not a limitation of \toolname{}, but an inherent property of the specification refinement problem itself because the problem is \emph{undecidable}. This is because, even for the simplest case of reachability specifications (e.g., “reach region $G$ with probability at least $p$”), deciding whether a given policy satisfies a specification is \emph{undecidable} for general continuous-state MDPs or probabilistic programs capable of encoding Turing-complete behavior~\cite{kaminski2015hardness}. Consequently, no specification refinement algorithm can be both sound and complete. \toolname{} therefore focuses on soundness, which is important towards ensuring that a policy for the refined task also solves the task defined by the original specification.


\section{Experimental Evaluation}
\label{sec:experiments_main}

We evaluate \toolname{} on its ability to diagnose and repair specification failures that prevent existing algorithms from learning satisfactory policies. Our experiments address three questions: (1) Can \toolname{} correctly identify which refinement type is needed for different failure modes? (2) Do the refinements enable learning from previously unlearnable specifications? (3) What are the requirements and limitations of the refinement process?
\subsection{Experimental Setup}

We integrate \toolname{} with two specification-guided RL algorithms: \textsc{DiRL}~\cite{dirl}, which uses Dijkstra-style graph search with systematic exploration, and \textsc{LSTS}~\cite{lsts}, which uses multi-armed bandits for edge selection with epsilon-greedy exploration. These algorithms differ fundamentally in their exploration strategies, allowing us to examine how \toolname{}'s effectiveness depends on the underlying learning algorithm. We evaluate on two domains specifically chosen to stress-test different aspects of specification refinement:

\textbf{n-Rooms:} Grid-based navigation with walls and doors, providing controlled tests of specific failure modes. State space: $(x, y, \theta, d) \in \mathbb{R}^4$ (position, angle to goal, distance). Action space: $(v, \theta) \in \mathbb{R}^2$ (velocity, direction). The n-rooms domain has been extensively used in specification-guided RL research~\cite{dirl, spectrl, vzikelic2024compositional} as it provides clear geometric structure while still presenting challenging long-horizon tasks. Its modular room structure naturally creates the types of specification failures we aim to address: trap states at room boundaries, dangerous narrow passages between rooms, and multiple alternative paths through different door configurations.

\textbf{PandaGym~\cite{gallouedec2021pandagym}:} Robotic manipulation requiring 3D navigation around obstacles. This domain tests refinement in high-dimensional continuous control where geometric intuitions may not apply directly. Following recent work showing the challenges of specification-guided RL in manipulation tasks~\cite{lsts}, we use this domain to validate that our convex hull and hyperplane-based refinements remain effective in high-dimensional spaces where human intuition about specification failures is limited.

For learning edge policies, both algorithms use PPO~\cite{DBLP:journals/corr/SchulmanWDRK17} with stable-baselines3~\cite{stable-baselines3} implementation, following the standard practice in recent specification-guided RL work~\cite{dirl, vzikelic2024compositional}. We use 2-layer networks (64 neurons each), learning rate 0.0003, and standard PPO hyperparameters.
In all experiments we evaluate refinements using a deliberately high satisfaction threshold ($p = 0.99$). The purpose of this choice is methodological: by selecting a probability level that is difficult to achieve under coarse or under-specified predicates, we can clearly observe how the cumulative probability of satisfying the specification improves as AutoSpec performs successive refinements. Using such a stringent threshold ensures that even small improvements in guidance become visible in the satisfaction curves and allows us to measure the full extent of the benefit provided by refinement, independent of how poorly the initial specification performs.

All experiments are repeated over five random seeds. Plots report the mean across the five runs, with error bars showing the empirical mean~$\pm$~variance. Specifically, each data point in the learning curves represents the performance of a policy trained for the distinct number of timesteps indicated on the x-axis (e.g., policies are trained for 80,000, 100,000, and 120,000 steps independently across 5 seeds). To estimate the specification satisfaction probability (y-axis), the trained policy for each edge is evaluated over 1000 rollout trajectories to empirically count successful versus failed attempts. The final success probabilities displayed in the plots are calculated using the product of success probability of the best path from start to goal. 

\subsection{Algorithm-Dependent Effectiveness: DiRL vs LSTS}

\begin{figure*}[!htb]
    \centering
        \includegraphics[width = 0.6\textwidth]{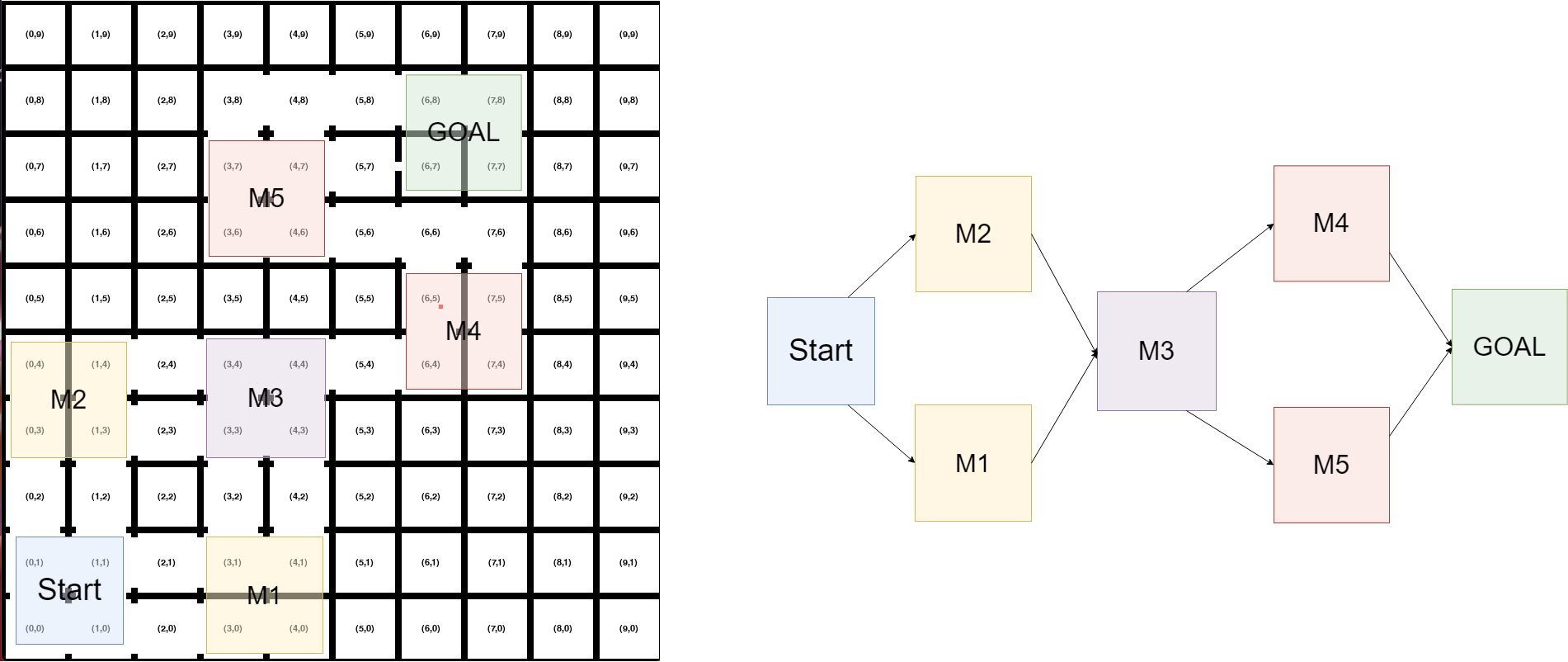}
        \caption{100-rooms Environment with marked regions its DAG specification}
    
    \label{fig:big_spec_env}
\end{figure*}
    
\begin{figure*}
\centering
    \begin{subfigure}{0.3\textwidth}
        \includegraphics[width=\textwidth]{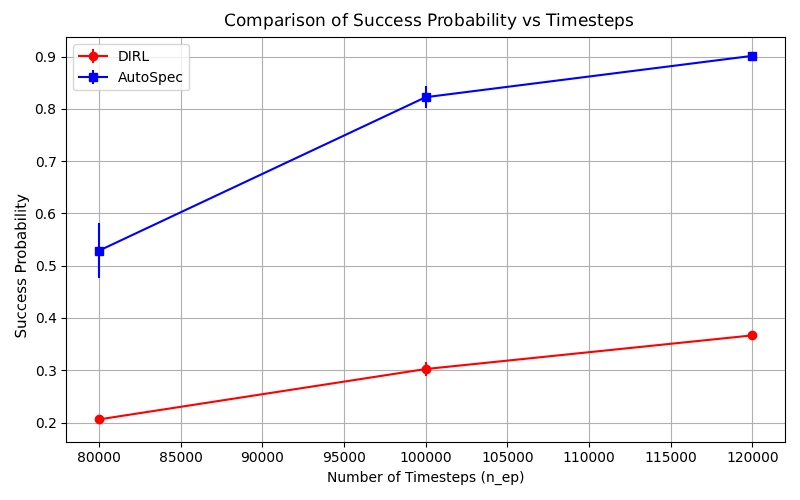} 
        \caption{Mid-goal DIRL performance}
    \end{subfigure}
    \begin{subfigure}{0.3\textwidth}
        \includegraphics[width=\textwidth]{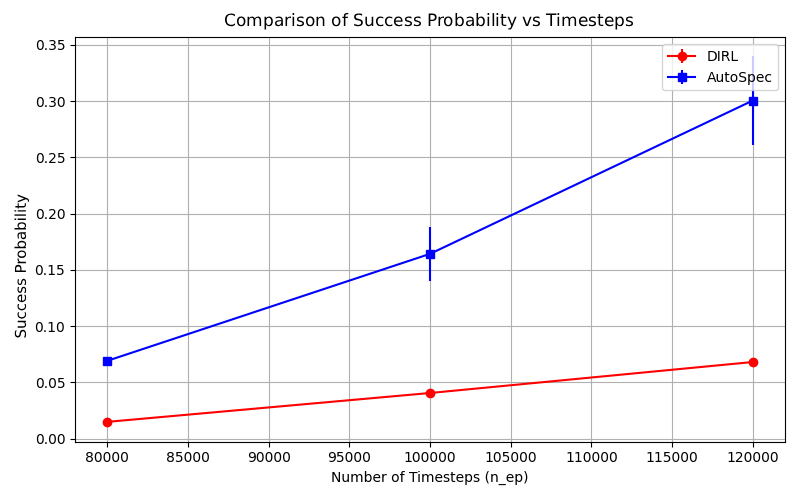}
        \caption{Full-spec DIRL performance}
    \end{subfigure}
    \begin{subfigure}{0.3\textwidth}
        \includegraphics[width=\textwidth]{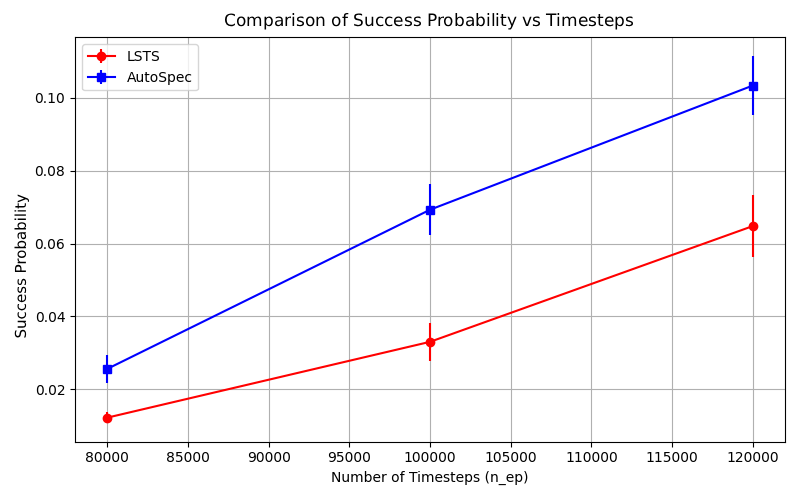}
        \caption{Mid-goal LSTS performance}
    \end{subfigure}
    
    \caption{Task satisfiability curves representing performances of DIRL and LSTS for sub-specifications and complete specification}
    \label{fig:big_spec}
\end{figure*}

Our experiments reveal that \toolname{}'s effectiveness depends critically on the base algorithm's exploration strategy. We demonstrate this through a 100-rooms environment (Figure~\ref{fig:big_spec_env}) with the complex specification:
$\phi = \phi_{start};(\phi_{m1} \text{ or } \phi_{m2});\phi_{m3}; (\phi_{m4} \text{ or } \phi_{m5});\phi_{goal}$

This specification structure, with multiple disjunctive branches and sequential compositions, represents the type of complex task decomposition that prior work~\cite{spectrl, dirl} has identified as necessary for real-world applications but challenging for existing algorithms. The 100-rooms scale specifically tests whether refinements remain effective when the state space is large enough that exhaustive exploration is infeasible, reflecting concerns raised in~\cite{lsts} about scalability of compositional methods.

\textbf{With DiRL (Successful Refinement).}
As shown in Figure~\ref{fig:big_spec}(a-b), DiRL's systematic exploration enables successful refinement. The algorithm explores edges in order of estimated difficulty, providing sufficient trajectory data for each edge before moving to the next. \toolname{} successfully applies ReachRefine on the $\phi_{m1}$ edge to remove unreachable portions of the target region, PastRefine on the $\phi_{m3}$ edge to identify successful starting regions, and OrRefine when direct paths fail to find alternative routes through $\phi_{m2}$. The satisfaction probability improves from near 0\% to approximately 60\% through these refinements.

\textbf{With LSTS (Refinement Failure).}
Figure~\ref{fig:big_spec}(c) shows LSTS failing on the same specification. The bandit-based exploration spreads effort across all edges simultaneously, preventing deep exploration of any single edge. Consequently, edges to M4, M5, and Goal achieve 0\% satisfaction, providing no successful trajectories for refinement computation. \toolname{} correctly reports its inability to refine without samples, demonstrating that refinement quality fundamentally depends on the base algorithm's exploration strategy.

\textbf{Evaluation on Randomized 100-rooms and predicate placement} To evaluate the generalization capabilities of our framework, we deployed \textsc{AutoSpec} in a procedurally generated 4-Way Gridworld where wall connectivity, and predicate placement were fully randomized for each seed (see the Appendix for generation details). This setup specifically tests the system's ability to synthesize policies without reliance on hand-engineered specifications or environment-specific heuristics. As shown in Figure~\ref{fig:experiment_overview}, \textsc{AutoSpec} significantly outperforms the \textsc{DIRL} baseline, achieving a terminal success probability of approximately $60\%$ compared to the baseline's stagnation at $20\%$. These results confirm that \textsc{AutoSpec} autonomously identifies and resolves task bottlenecks, raising success rate of critical transitions from $<20\%$ to $>90\%$ via automatic refinement, see the Appendix.

\vspace{-5pt}
\subsection{Evaluation of Individual Refinements}

We design targeted experiments isolating specific failure modes to validate each refinement procedure.

\textbf{SeqRefine: Trap State Elimination (Figure~\ref{fig:results_reach}).}
\textbf{Setup:} 9-rooms environment where the goal region includes a blocked room creating a trap state.
\textbf{Failure mode:} Agent reaches the trap portion of the goal and cannot escape.
\textbf{Refinement:} ReachRefine identifies that successful trajectories only reach the accessible portion of the goal. The refined specification excludes the trap region: $b_r = b \cap \text{ConvexHull}(\text{reached states})$.
\textbf{Result:} Satisfaction probability improves from 15\% to 85\%, demonstrating \toolname{}'s ability to learn environmental constraints not captured in the original specification.

\textbf{SeqRefine: Safety Constraint Discovery (Figure~\ref{fig:results_avoid}).}
\textbf{Setup:} 9-rooms with a narrow dangerous passage below the goal.
\textbf{Failure mode:} Shortest path goes through narrow passage where agent frequently fails.
\textbf{Refinement:} AvoidRefine identifies failure states near the narrow passage. The refined specification expands the avoid region: $c_r = c \setminus \text{ConvexHull}(\text{last 10 failure states})$.
\textbf{Result:} Agent learns to use wider but longer safe path, improving satisfaction from 30\% to 75\%.

\textbf{AddRefine: Waypoint Introduction (Figure~\ref{fig:results_add}).}
\textbf{Setup:} Long-horizon navigation across multiple rooms.
\textbf{Failure mode:} Direct path too complex for single policy to learn reliably.
\textbf{Refinement:} AddRefine identifies midpoints of successful trajectories and introduces intermediate vertex $u''$.
\textbf{Result:} Decomposes task into two manageable subtasks, improving satisfaction from 20\% to 90\%.

\textbf{PastRefine: Initial State Partitioning (Figure~\ref{fig:results_reach_past}).}
\textbf{Setup:} Starting region includes states from which goal is unreachable.
\textbf{Failure mode:} Policy cannot succeed from certain initial states.
\textbf{Refinement:} PastRefine learns hyperplane separating successful from failing starts.
\textbf{Result:} Focuses learning on viable initial states, improving satisfaction from 40\% to 80\%.

\textbf{OrRefine: Alternative Path Discovery (Figure~\ref{fig:results_or}).}
\textbf{Setup:} Specification with multiple possible paths: $\phi_{MID1};\phi_{GOAL}$ or $\phi_{MID2};\phi_{GOAL}$.
\textbf{Failure mode:} Direct path through MID1 blocked.
\textbf{Refinement:} OrRefine adds edge $\phi_{MID1} \rightarrow \phi_{MID2}$, creating alternative route.
\textbf{Result:} Enables satisfaction through alternate path when direct path has 0\% success.

\subsection{High-Dimensional Validation: PandaGym}
\begin{figure}[!htb]
    \centering
    \subfloat{{\includegraphics[width=0.23\linewidth]{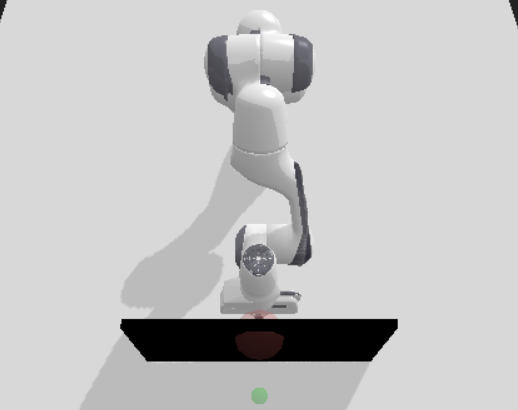} }}
    \subfloat{{\includegraphics[width=0.23\linewidth]{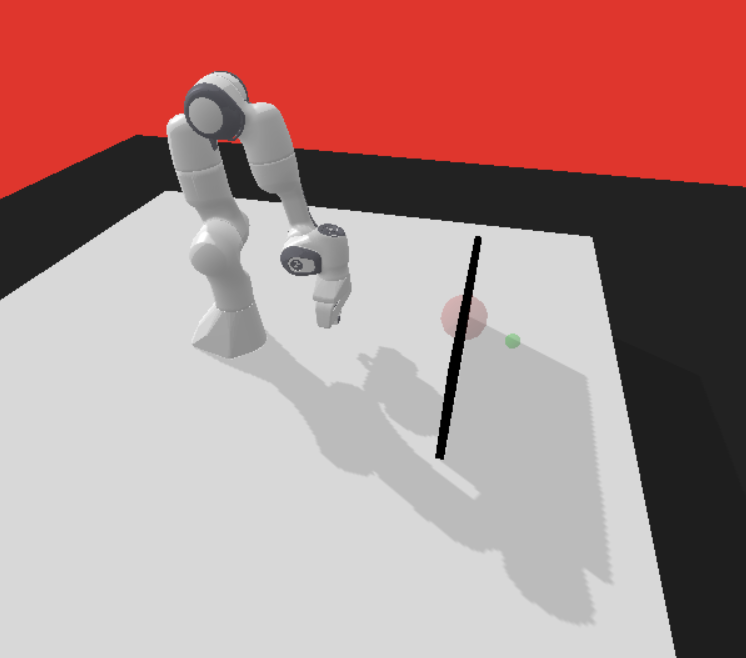} }}
    \subfloat{{\includegraphics[width=0.23\linewidth]{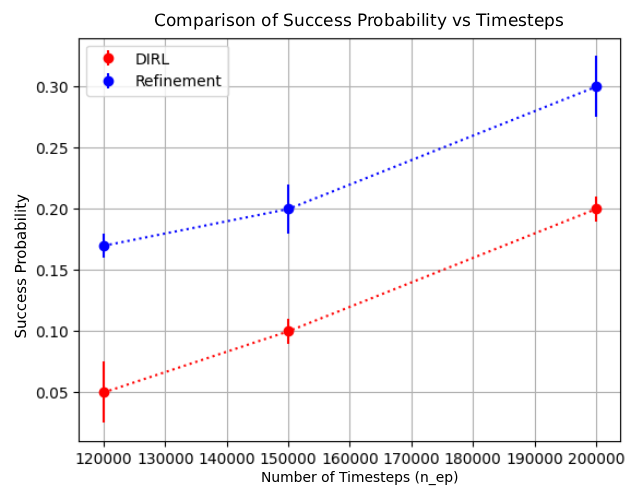} }}
    \subfloat{{\includegraphics[width=0.23\linewidth]{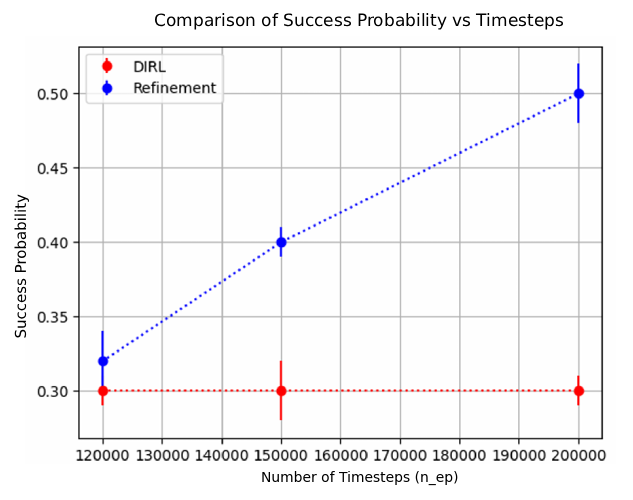} }}
    \caption{Evaluation of \toolname{} on PandaGym: (a) Two perspectives of the environment (1st and 2nd Figures), where the red region is an intermediate goal and an invisible wall blocks direct paths. (b) Performance of DiRL with and without \toolname{}: ReachRefine on first edge (3rd Figure) and PastRefine on second edge (4th Figure).}
    \label{fig:pandagym}
\end{figure}

To validate beyond grid environments, we test \toolname{} on PandaGym's continuous 3D manipulation task. The specification requires navigating around an invisible wall: $(reach \text{ red-region avoid wall}); (reach \text{ green-region avoid wall})$. The invisible wall creates a challenging scenario where the agent cannot directly observe the obstacle, making specification refinement crucial.

As shown in Figure~\ref{fig:pandagym}, \toolname{} with DiRL successfully applies ReachRefine on the first edge to identify and exclude unreachable portions of the red region behind the wall, focusing the policy on achievable subgoals. On the second edge, PastRefine learns that only certain approach angles from the red region lead to successful reaching of the green region, effectively partitioning the intermediate state space based on trajectory outcomes.  This demonstrates that \toolname{}'s geometric refinements (convex hulls for ReachRefine, hyperplanes for PastRefine) remain effective in high-dimensional spaces where human intuition about the specification failures would be difficult. The success in this domain is particularly noteworthy because the refinements must capture 3D spatial relationships without explicit knowledge of the obstacle geometry.

\textbf{Computational Overhead.} 
\textsc{AutoSpec} avoids full retraining by only updating the policies associated with the identified subset of refined edges $\mathcal{R}$. The total computational cost is formalized as $T_{\mathrm{total}} = T_{\mathrm{base}} + \sum_{e\in\mathcal{R}} T_e$. Since $|\mathcal{R}|$ is typically small relative to the initial graph size, the aggregate overhead is bounded (empirically $T_{\mathrm{total}} \leq 2 T_{\mathrm{base}}$). In the 100-room experiments, the baseline required $\sim 240$s to evaluate the 8 fixed edges, while \textsc{AutoSpec} averaged $390 \pm 42$s. This overhead corresponds directly to the training of 4--7 additional refinement edges per seed, with the observed variance ($\sigma \approx 42\text{s}$) driven by the differing topological complexity of the randomized environments. This computational investment is highly efficient, scaling linearly with the number of detected bottlenecks rather than the global state space size. Given the substantial improvement in success probability (from $\approx 20\%$ to $\approx 60\%$), this bounded overhead represents a favorable trade-off for achieving robust autonomy in stochastic domains.

\vspace{-5pt}
\section{Conclusion}
\label{sec:conclusion}

We presented \toolname{}, a framework for automated refinement of coarse-grained logical specifications in reinforcement learning. \toolname{} addresses two common specification issues — coarse formulas and coarse labeling functions through four refinement procedures that maintain formal soundness. Our experiments on n-rooms and PandaGym environments demonstrate that \toolname{} can improve specification satisfiability when integrated with existing algorithms like DiRL and LSTS.

Our evaluation also reveals fundamental limitations: \toolname{} requires sufficient exploration data from the base algorithm to compute meaningful refinements. When algorithms fail to generate successful trajectories (as LSTS did on complex specifications), refinement becomes impossible. Despite these limitations, \toolname{} represents the first systematic approach to automatically refining logical specifications based on learning failures. Future work should address reducing exploration requirements for refinement and extending beyond SpectRL to more expressive temporal logics, such as infinite-horizon $\omega$-regular specifications. While \toolname{} currently relies on finite witnesses, it could be adapted to these settings by decomposing tasks into a finite prefix (amenable to our current DAG-based refinement) and a cyclic suffix (which would require extending our witness analysis to handle infinite behaviors). The design of good specifications remains challenging in practice, and automated refinement is an important step toward making specification-guided RL more practical.


\bibliography{refs}
\bibliographystyle{iclr2026_conference}

\clearpage
\appendix

\section{Appendix / supplemental material}

\subsection{Refinement Algorithms}

Here we present pseudo-code for the individual refinement algorithms described in Section~\ref{subsec:refine}

\begin{algorithm}[!h]
\caption{ReachRefine}\label{alg:reachmod}
\begin{algorithmic}
\REQUIRE $b:=\beta(u'), \zeta$
\STATE $S_r \gets \{s  \mid s \in b \cap \zeta\}$\; \textit{Collect all the goal region states from the trajectories}
\STATE $b_r \gets b \cap \text{ConvexHull}(S_r)$\; \textit{Create the convex hull of the collected states}
\RETURN $b_r$
\end{algorithmic}
\end{algorithm}

\begin{algorithm}[!h]
\caption{AvoidRefine}\label{alg:avoidmod}

\begin{algorithmic}
\REQUIRE $c:= \beta(e), \zeta$
\STATE $O_r \gets \{\}$\; \textit{Initialize new avoid region with an empty set}
\FOR{$\zeta_i$ \textbf{in} $\zeta$}
\IF{$\zeta_i[-1] \not\in c$ }
\STATE $O_r \gets O_r \cup \{s_j \mid s_j \in \zeta_i \bigwedge (\textit{len}(\zeta_i)-j)\leq k \}$

\textit{Append the last $k$ states from every trajectory that ended up in the avoid region}
\ENDIF
\ENDFOR
\STATE $c_r \gets c \backslash \text{ConvexHull}(O_r)$\; \textit{Create a convex hull around the collected states and remove it form the original safe region }
\RETURN $c_r$
\end{algorithmic}
\end{algorithm}

\begin{algorithm}[!h]
\caption{SeqRefine: Refining Edge $e = u \rightarrow u'$}\label{alg:seqmod}
\begin{algorithmic}

\REQUIRE Edge $e = u \rightarrow u'$, Graph $G$, set of trajectories $\zeta$. 
\STATE $b = \beta(u')$ \textit{States in the reach predicate}
\STATE $c = \beta(e)$ \textit{States in the avoid predicate}
\STATE $b_r \gets \textit{ReachRefine}(b, \zeta)$
\STATE $c_r \gets \textit{AvoidRefine}(c, \zeta)$
\STATE $u_r \gets [\beta(u_r) = b_r]$ \textit{Redefine target node with new predicate}
\STATE $e_r \gets [u \rightarrow u_r, \textit{with } \beta(e_r) = c_r] $ \textit{Redefine edge with new predicate}
\STATE $G' \gets G\setminus[e\gets e_r] $ \textit{Replace edge with refinement}

\RETURN $G'$
\end{algorithmic}
\end{algorithm}

\begin{algorithm}[!h]
\caption{AddRefine}\label{alg:addmod}
\begin{algorithmic}
\REQUIRE Edge $e = u \rightarrow u'$, Graph $G$, set of trajectories $\zeta$. 
\STATE $S_r \gets \{\}$\;
\FOR{$\zeta_i$ \textbf{in} $\zeta$}
\IF{$\zeta_i \models e$ [i.e. trajectory was successful]}
    \STATE $S_r \gets S_r \cup \zeta_i[len(\zeta_i)//2]$ \textit{Add center of trajectory as waypoint}
\ENDIF

\ENDFOR

\STATE $b_r \gets \textit{ConvexHull}(S_r) \cap \beta(e)$
\STATE $u'' \gets [\beta(u'') = b_r]$ \textit{Define target node for waypoint}
\STATE $e'' \gets [u \rightarrow u'', \textit{with } \beta(e'') := \beta(e)] $ 
\STATE $e' \gets [u'' \rightarrow u', \textit{with } \beta(e') := \beta(e)] $ 
\textit{Define edges with new waypoint predicate}

 \STATE $G' \gets G\setminus[e\gets [e'';e']] $ \textit{Replace edge $e$ with composition of new edges}

\RETURN $G'$

\end{algorithmic}
\end{algorithm}

\begin{algorithm}[!h]
\caption{PastRefine: Refining Abstract Graph Exploration}\label{alg:graph_refinement}
\begin{algorithmic}
\REQUIRE Edge $e = u \rightarrow u'$, Graph $G$, set of trajectories $\zeta$. 

\STATE $S \gets \{\}$\; $S_r \gets \{\}$\;
\FOR{$\zeta_i$ \textbf{in} $\zeta$}
\STATE $S \gets S \cup \zeta_i[0]$ \textit{Collect start states from all trajectories}
\IF{$\zeta_i \models e$ }
    \STATE $S_r \gets S_r \cup \zeta_i[0]$ \textit{Collect start states from successful  trajectories}

\ENDIF

\ENDFOR
\STATE Identify a hyperplane $H$ separating the $S_r$ and $S\setminus S_r$
\STATE $b_r \gets \{s \in S : H(s) \geq 0\}$
\STATE $u^* \gets [\beta(u^*) = b_r]$ \textit{Redefine initial node with new predicate}
\STATE $e_r \gets [u^* \rightarrow u', \textit{with } \beta(e_r) := \beta(e)] $ \textit{Redefine edge with new predicate}
\STATE $G' \gets G\setminus[e\gets e_r] $ \textit{Replace edge with refinement}
\RETURN $G'$
\end{algorithmic}
\end{algorithm}

\begin{algorithm}[!h]
\caption{OrRefine: Disjunctive Specification Refinement}
\label{alg:dis_spec_refinement}
\begin{algorithmic}

\REQUIRE Edge $e = u \rightarrow u'$, Graph $G$, set of trajectories $\zeta$. 


        
\STATE $E = \{e_i \in G \mid e_i = u_i \rightarrow u' \} $
\textit{Collect all 'parents' of $u'$}

\FOR{$e_i \in E$}
    \STATE $e_{new} \gets [u \rightarrow u_i, \textit{with } \beta(e_{new}) := \beta(e) \text{ and } \beta(e_i) := \beta(e) \cap\beta(e_{i})] $ \textit{Define edges from source to parents}
    \STATE $G \gets G\cup[e_{ui}] $
    \textit{Add new edge to graph}
\ENDFOR


\RETURN G
\end{algorithmic}
\end{algorithm}

\subsection{Proof of Theorem 1}
\label{sec:fullproof}

\textbf{Theorem 1} (Correctness of \textsc{AutoSpec})
\textit{Given an abstract graph $G$ of a SpectRL specification $\phi$ and an edge $e$, \textsc{AutoSpec} computes a specification $\phi_r$ with abstract graph $G_r$ such that $\phi_r$ refines $\phi$. That is, for any MDP trajectory $\zeta$, we have $(\zeta \models \phi_r) \implies (\zeta \models \phi)$.}

\smallskip\noindent{\em Proof.} To prove the theorem, it suffices to show that for each of the four refinement subprocedures, if they return an abstract graph $G_r$, then the corresponding specification $\phi_r$ is a refinement of the input specification $\phi$.

By the definition of abstract reachability, we have $(\zeta \models \phi) \Leftrightarrow (\zeta \models G)$ and $(\zeta \models \phi_r) \Leftrightarrow (\zeta \models G_r)$. Hence, to prove that $(\zeta \models \phi_r) \Rightarrow (\zeta \models \phi)$ which is the definition of specification refinement as in Definition~\ref{def:refinement}, it suffices to prove that $(\zeta \models G_r) \Rightarrow (\zeta \models G)$. We prove this claim for each refinement subprocedure.

{\bf SeqRefine.} Suppose that $G_r = \textsc{SeqRefine}(e,G,\zeta)$. Let $e = u \rightarrow u'$. By our design of SeqRefine, the abstract graph $G_r$ has the same vertex set, edge set and labeling function as $G$, with the only difference being that $\beta_r(u') \subseteq \beta(u')$ due to ReachRefine and $\beta_r(e) \subseteq \beta(e)$ due to AvoidRefine. Hence, every trajectory $\zeta$ that satisfies all reach-avoid tasks in $G_r$ must also satisfy those in $G$, giving us $(\zeta \models G_r) \Rightarrow (\zeta \models G)$.

{\bf AddRefine.} Suppose that $G_r = \textsc{AddRefine}(e,G,\zeta)$. Let $e = u \rightarrow u'$. AddRefine introduces a new vertex $u''$ and replaces edge $e$ with two sequentially composed edges $e'' = u \rightarrow u''$ and $e' = u'' \rightarrow u'$ where $\beta_r(e'') = \beta_r(e') = \beta(e)$. Any trajectory satisfying the refined path through $u''$ must visit the intermediate waypoint while respecting the original safety constraints, thus also satisfying the original edge specification. Therefore, $(\zeta \models G_r) \Rightarrow (\zeta \models G)$.

{\bf PastRefine.} PastRefine refines the region associated to vertex $u$ by restricting it to $\beta_r(u) \subseteq \beta(u)$. This refinement affects both edge $e = u \rightarrow u'$ and all edges incoming to $u$. Since the refined region is a subset of the original, any trajectory satisfying the refined specification must originate from states that were valid in the original specification. Hence, $(\zeta \models G_r) \Rightarrow (\zeta \models G)$.

{\bf OrRefine.} Suppose that $G_r = \textsc{OrRefine}(e,G,\zeta)$. OrRefine only adds edges between existing vertices in $G$. Specifically, for a problematic edge $e = u \rightarrow u'$, it identifies existing edges $e_i = u_i \rightarrow u'$ and adds new edges $e_{new} = u \rightarrow u_i$ where $\beta(e_{new}) = \beta(e)$ and $\beta(e_i) = \beta(e) \cap\beta(e_{i})$. 

Consider a trajectory $\zeta$ that satisfies $G_r$ via a newly added path $u \rightarrow u_i \rightarrow u'$. Since:
(1) Both $u_i$ and $u'$ existed in the original vertex set of $G$,
(2) The edge $u_i \rightarrow u'$ existed in the original edge set of $G$,
(3) The new edge $u \rightarrow u_i$ maintains the safety constraints of the original edge ($\beta(e_{new}) = \beta(e)$),

the trajectory $\zeta$ reaches $u'$ through a combination of transitions that respect all original safety constraints and only uses vertices from the original specification. The path through $u_i$ represents a valid alternative route in the original specification structure. Therefore, $(\zeta \models G_r) \Rightarrow (\zeta \models G)$.

Thus, all four refinement procedures preserve specification soundness. \hfill\qed

\subsection{Experiments}
\label{sec:experiments}

\subsubsection{Procedural Environment Generation}
\label{sec:procedural_environment_generation}
To evaluate the robustness of the proposed refinement framework, we utilize a procedurally generated Continuous Gridworld environment (Figure~\ref{fig:experiment_overview}. The environment layout and region locations are randomized for each experimental seed to ensure that the agent learns generalized navigation skills rather than memorizing a specific map layout. We use the same topological layout as in Figure~\ref{fig:big_spec_env}. The generation process follows a three-step strategy:

\subsubsection{Grid and Region Definitions}
The domain is defined as a grid of $10 \times 10$ rooms, where each room has a spatial dimension of $8.0 \times 8.0$ units. We define a set of logical regions $\mathcal{R} = \{Start, Goal, M_1, M_2, MG, MG_1, MG_2\}$. Each region $r \in \mathcal{R}$ is physically instantiated as a $2 \times 2$ block of contiguous rooms.

\subsubsection{Randomized Placement Strategy}
The placement of regions is performed stochastically to maximize variability while maintaining a solvable structure:
\begin{enumerate}
    \item \textbf{Start and Goal Initialization:} We define the four corners of the grid as candidate zones. To ensure traversal complexity, the $Start$ and $Goal$ regions are restricted to \textit{opposite corners}. A pair of opposite corner zones (e.g., Top-Left and Bottom-Right) is selected uniformly at random. The $Start$ region is assigned to one zone and the $Goal$ to the other, denoted as:
    \begin{equation}
        (Pos_{start}, Pos_{goal}) \sim \text{Uniform}(\{(C_{TL}, C_{BR}), (C_{TR}, C_{BL})\})
    \end{equation}
    where $C$ represents the set of valid indices for a $2 \times 2$ block in a specific corner.
    
    \item \textbf{Intermediate Region Placement:} The remaining intermediate regions (e.g., $M_1, M_2, \dots$) are placed randomly within the grid. Their locations are sampled uniformly from the set of all valid $2 \times 2$ block coordinates, subject to the constraint that no two regions may spatially overlap:
    \begin{equation}
        Pos_{r} \sim \text{Uniform}(\mathcal{G}_{valid} \setminus \bigcup_{r' \in \mathcal{R}_{placed}} Area(r'))
    \end{equation}
    This ensures that all sub-goals are distinct and distributed stochastically across the map.
\end{enumerate}

\subsubsection{Stochastic Connectivity}
Once regions are placed, the connectivity between adjacent rooms is generated probabilistically. For every pair of adjacent rooms $(u, v)$ in the grid, a connecting door is instantiated via a Bernoulli trial:
\begin{equation}
    P(\text{Door}_{u,v}) = 0.5
\end{equation}

\begin{figure}[htbp]
    \centering
    \begin{subfigure}[b]{0.37\textwidth}
        \centering
        \includegraphics[width=\textwidth]{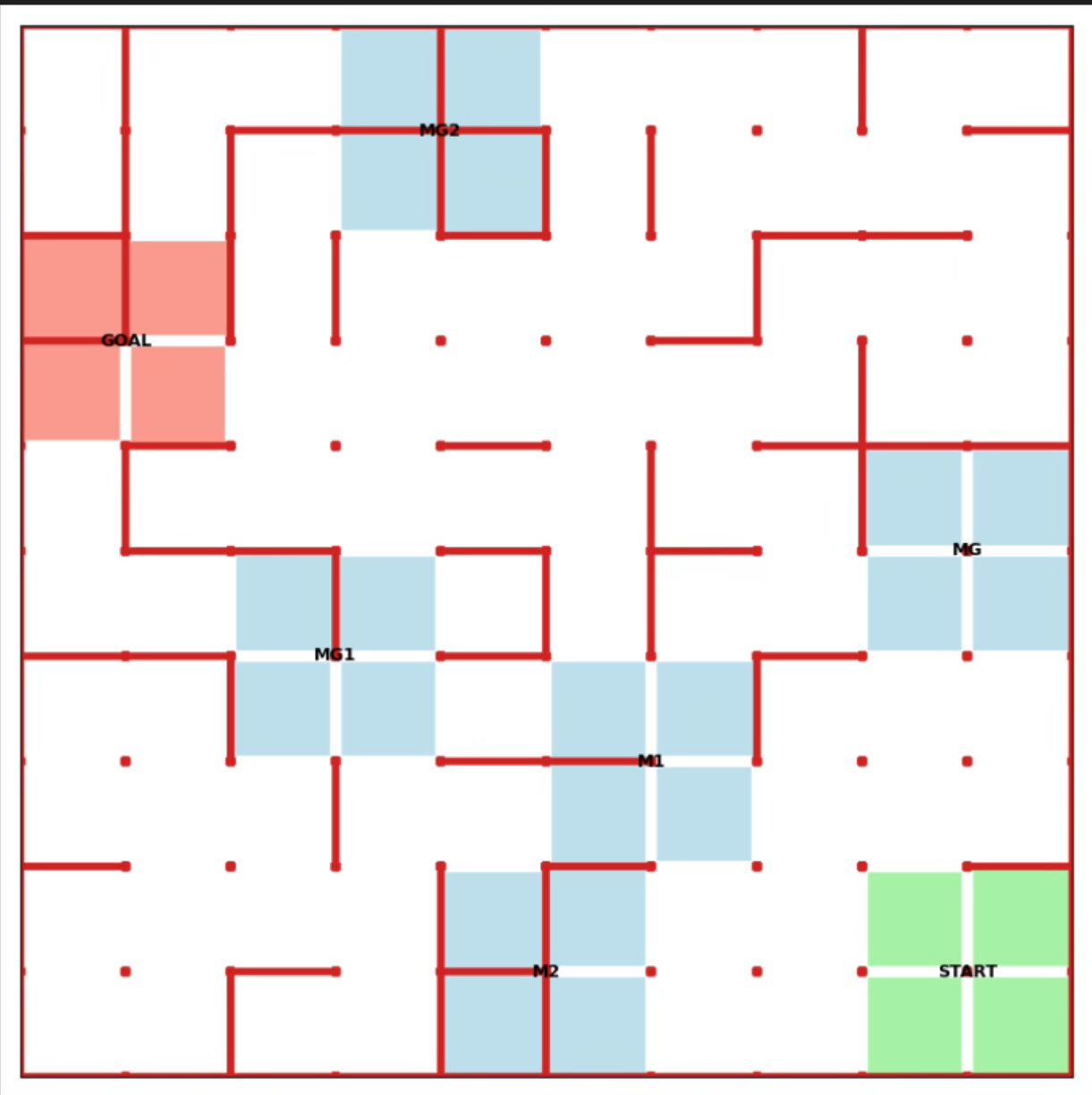}
        \caption{Randomized Environment and Predicates Instance}
        \label{fig:env_layout}
    \end{subfigure}
    \begin{subfigure}[b]{0.45\textwidth}
        \centering
        \includegraphics[width=\textwidth]{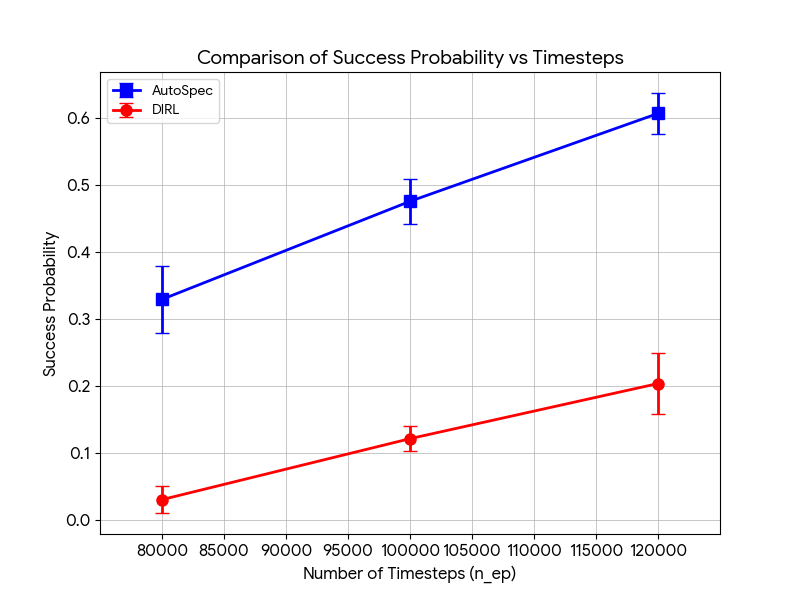}
        \caption{Success Probability Comparison}
        \label{fig:results_plot}
    \end{subfigure}
    
    \caption{\textbf{Experimental Results on the 100-room Gridworld.} 
    (a) A visualization of one of the procedurally generated environment instance, showing the randomized placement of the Start (green), Goal (red), and intermediate refinement regions (blue). 
    (b) Comparative performance analysis showing the mean success probability of the proposed AutoSpec method versus the DIRL baseline over 80k and 100k timesteps.}
    \label{fig:experiment_overview}
\end{figure}

\subsubsection{Results for Randomized Predicate Location}
\label{sec:results_random}

The experimental results on the randomized 100-room Gridworld demonstrate that \textsc{AutoSpec} operates effectively without the need for carefully crafted specifications or hand-engineered environments. As illustrated in Figure~\ref{fig:experiment_overview}, the environment introduces significant complexity through randomized wall configurations and arbitrarily placed predicate regions. We evaluated performance across 5 different procedurally generated environments (one per seed) and report the aggregated results in Figure~\ref{fig:results_plot} (b). Under these stochastic conditions, the \textsc{DIRL} baseline fails to synthesize a robust policy, stagnating at a low success probability.

In contrast, \textsc{AutoSpec} demonstrates superior adaptability by automatically refining the specification graph to overcome structural anomalies. While the global end-to-end success rate plateaus at approximately $50\%$, this limitation is not a failure of the refinement process itself, but rather the result of specific topological challenges inherent to the randomized domain. Our analysis of the transition logs reveals two specific phenomena that dictate performance:

\textbf{The "Bridge" Bottleneck.} 
In $80\%$ of the random seeds, the primary failure mode is the transition exiting the central convergence point ($MG \to MG_1$). While agents can reliably reach the $MG$ region ($>90\%$ success), the unrefined transition to the subsequent $Mid_1$ region frequently collapses to a success probability of $10$--$15\%$ due to randomized wall placements creating narrow or non-linear passages. \textsc{AutoSpec} addresses this by applying \textit{AddRefine} to introduce intermediate waypoints ($MG \to MG_{add} \to MG_1$) or \textit{ReachRefine} to tighten target constraints. Empirical logs show these refinements consistently raise the local success rate of this specific bottleneck edge to $50$--$70\%$, thereby recovering end-to-end traversability where standard methods fail.

\textbf{Autonomous Corridor Switching.} 
A critical advantage of the refinement process is the ability to dynamically switch logical corridors. The global task allows for alternate paths ($Start \to \{M_1, M_2\} \to MG \to \{MG_1, MG_2\} \to Goal$). In instances where the path through $MG_1$ remains stochastically blocked (success $<5\%$) even after refinement attempts, \textsc{AutoSpec} effectively prunes this branch. The search strategy then shifts probability mass to the alternative $MG_2$ branch. In our experiments, we observed seeds where the unrefined agent effectively "gave up" due to the difficulty of $MG_1$, whereas the refined agent successfully discovered and optimized the alternative $MG \to MG_2$ route ($>78\%$ local success), proving that automated refinement acts as a form of robust structural exploration.

\subsubsection{Results for Individual Refinements}

\begin{figure}[h]%
    \centering
    \subfloat{{\includegraphics[width=0.3\linewidth]{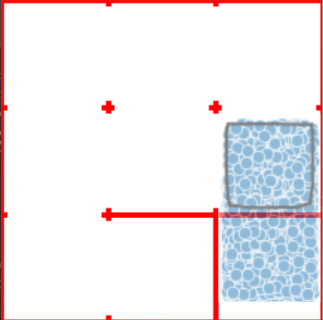} }}%
    \subfloat{{\includegraphics[width=0.4\linewidth]{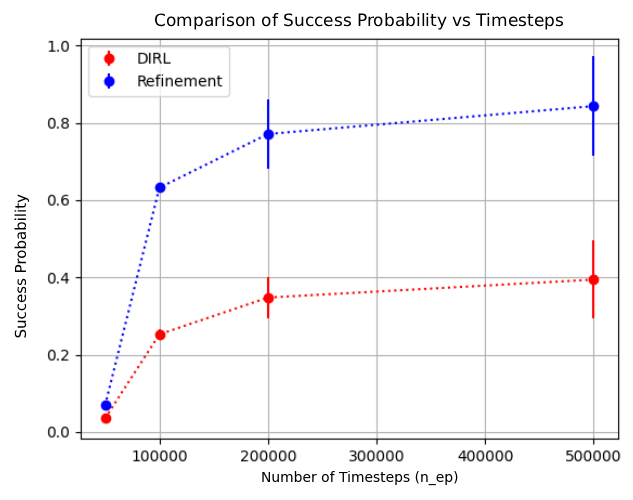} }}%
    \qquad
    
    \caption{Evaluation of Reach Probabilities in the 9-Rooms Environment. (a) The layout of the 9-rooms environment, showing the walls, doors, and goal regions, and the estimated convex hull for the new reach region, showing how the refinement process effectively restructs the reachable states, leading to better satisfaction of the specification (b) A comparison of reach probabilities between DIRL~\cite{dirl} and the proposed AutoSpec approach. The x-axis denotes the number of steps, and the y-axis denotes the estimated probability of success.}
    
    \label{fig:results_reach}%
\end{figure}
\begin{figure}[!h]%
    \centering
    \subfloat{{\includegraphics[width=0.3\linewidth]{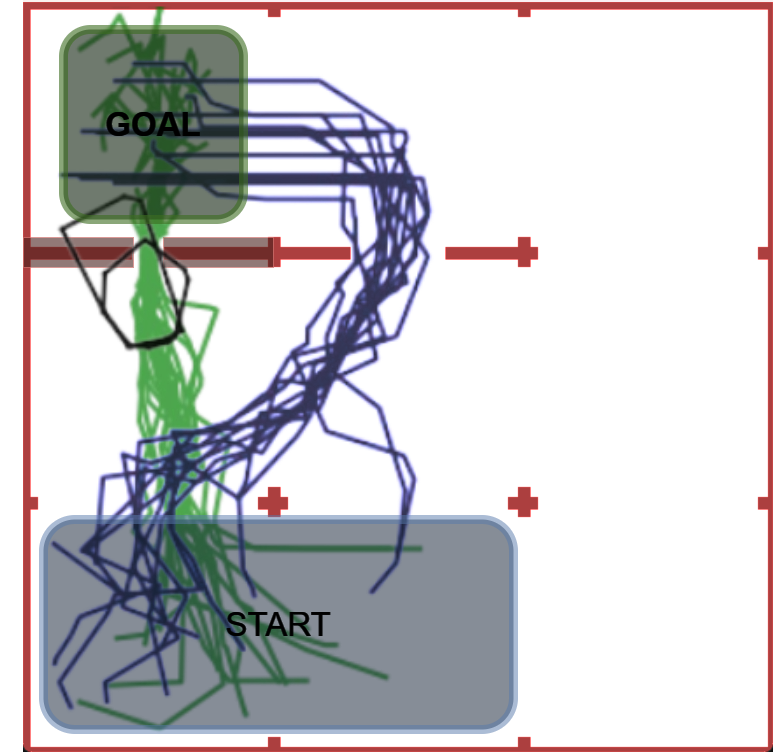} }}%
    \subfloat{{\includegraphics[width=0.4\linewidth]{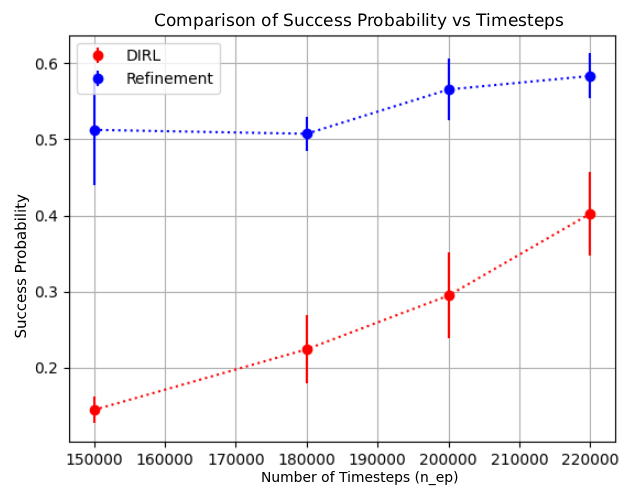} }}%
    
    \caption{Results of Avoid refinement. (a) The layout of the 9-rooms environment, showing the walls, doors, goal regions and avoid regions (red) and learned trajectories before (green) and after (blue) refinement, with new estimated avoid regions (black) (b) A comparison of reach probabilities between DiRL and the proposed Avoid Refinement. }
    
    \label{fig:results_avoid}%
\end{figure}

We conducted multiple experiments to validate our approach to refining coarse-grained SpectRL specifications for solving RL tasks.  
The goal of our experiments is to compare the performance of the original \textsc{DiRL}~\cite{dirl} algorithm and our integration of \textsc{DiRL} with \textsc{AutoSpec}, thus showcasing the ability of \textsc{AutoSpec} to refine SpectRL specifications that are challenging for the existing algorithms for RL from logical specifications. For learning edge policies, in both cases we use Proximal Policy Optimization (PPO) \cite{DBLP:journals/corr/SchulmanWDRK17}, implemented using stable-baselines3 \cite{stable-baselines3}. We employ a 2-layer neural network, each layer containing 64 neurons. We consider two environments.

\textbf{9 Rooms.} The 9 Rooms environment consists of walls blocking access to some rooms and doors allowing access to adjacent rooms. It has a 4-D continuous state space $(x, y, \theta, d) \in \mathbb{R}^4$, representing the 2D position, angle to the goal, and distance to the goal. We consider several SpectRL specifications which are translated into abstract graphs. The start position is sampled from the region associated to the source vertex, and the goal position is sampled from the region associated to the target vertex of the abstract graph. The 2-D continuous action space determines the velocity and direction of the agent $(v, \theta) \in \mathbb{R}^2$, with the new position calculated as $s' = s + (v \cos(\theta), v \sin(\theta))$.


\textbf{PandaGym.} The Pandagym \cite{gallouedec2021pandagym} reach environment has a robotic arm with an object picked up and the task is to place the object at the correct location. A wall blocking the path to the goal is invisible to the robot. The state space consists of the current position of the gripper arm in 3D and the goal position.

\begin{figure}[!h]%
    \centering
    \subfloat{{\includegraphics[width=0.3\linewidth]{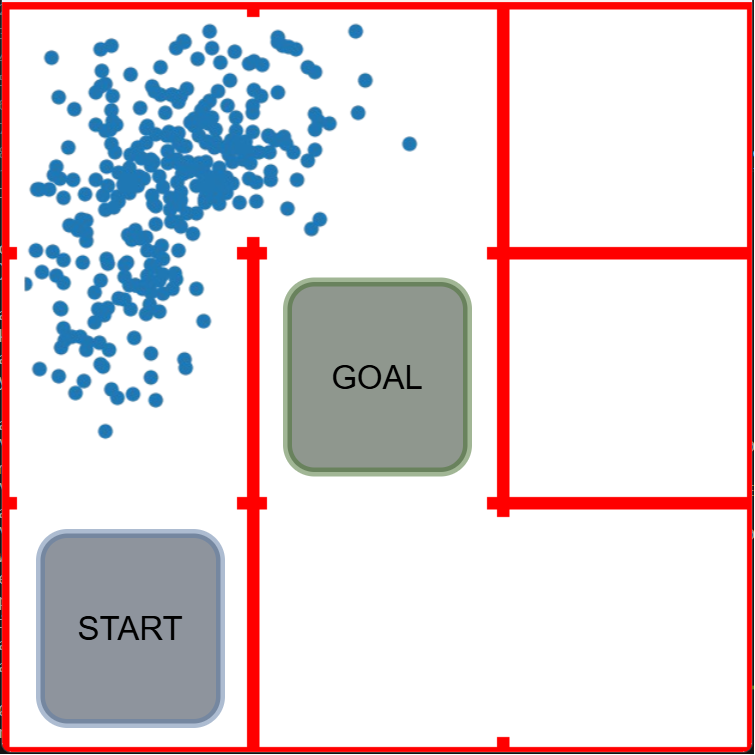} }}%
    \subfloat{{\includegraphics[width=0.4\linewidth]{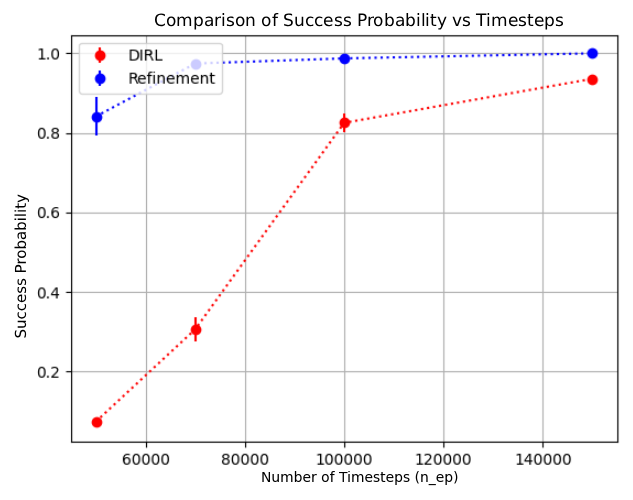} }}%
    
    \caption{Results of AddRefine in the 9-Rooms Environment. (a) The environment is annotated with start and goal regions (b) Learning curves comparing reach probabilities for DiRL and AutoSpec.}
    
    \label{fig:results_add}%
\end{figure}

\begin{figure}[!h]%
    \centering
    \subfloat{{\includegraphics[width=0.3\linewidth]{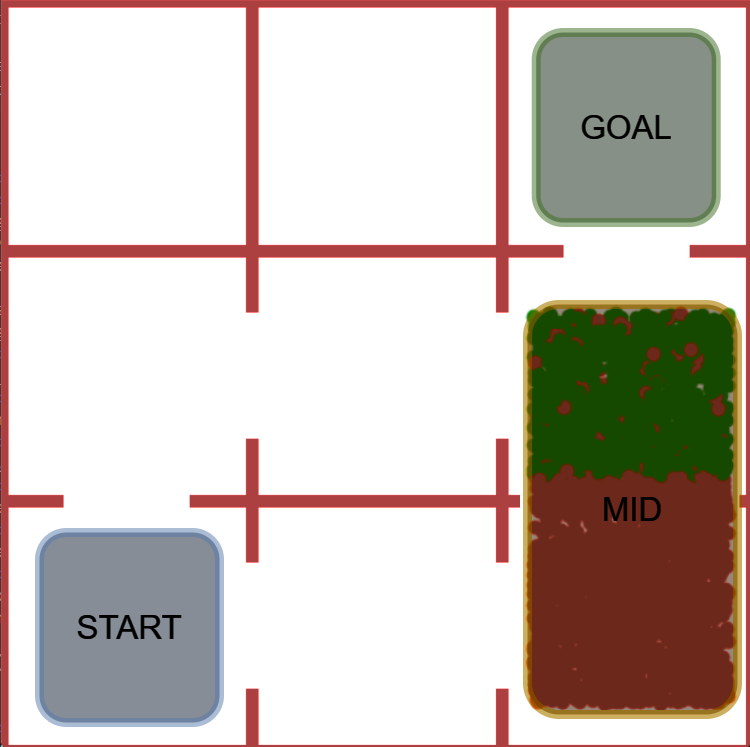} }}%
    \subfloat{{\includegraphics[width=0.4\linewidth]{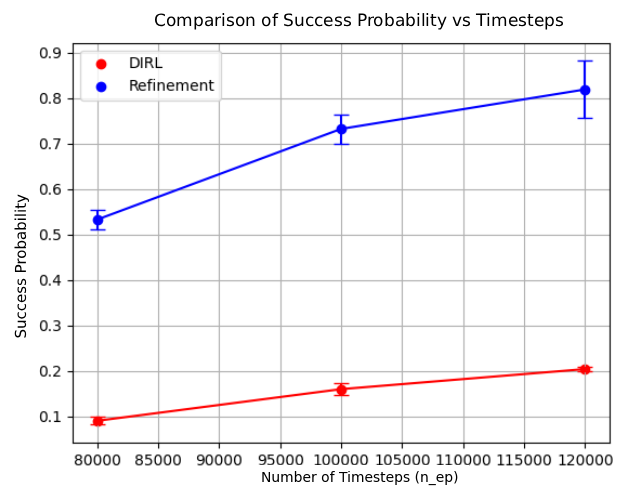} }}%
    
    \caption{Results of Sequential Specification Refinement in the 9-Rooms Environment. (a) The environment annotated with the distribution of states in the MID region from which the GOAL region can be reached. (b) Learning curves comparing reach probabilities for DiRL and AutoSpec.}
    
    \label{fig:results_reach_past}%
    
\end{figure}

{\bf Experiment 1: Atomic predicate refinement.} To illustrate how an incorrect specification is identified and corrected using Algorithm \ref{alg:reachmod}, we consider a 9 Rooms environment in Figure \ref{fig:results_reach}. In this environment, one room in the goal region is blocked, representing the incorrect specification. Figure \ref{fig:results_reach} displays the learning curves for both the original and refined specifications, demonstrating the performance improvements achieved through the refinement process. Algorithm \ref{alg:avoidmod} can be empirically verified by creating a 9 Rooms environment as shown in Figure \ref{fig:results_avoid}, where Figure \ref{fig:results_avoid} (a) shows the goal region along with avoid region (red). To improve probability of satisfaction the agent should avoid the narrow door below the goal and use the longer but safer route to approach the goal from the side. We see the learned trajectories before and after refinement in Figure \ref{fig:results_avoid} (a), where the new avoid region blocks the narrow door, effectively causing the agent to learn a policy that uses the wider door on the side. This helped improve specification satisfiability. 
\paragraph{Experiment 2: Sequential refinement.} We created a specification $\phi_{goal}$ to evaluate Algorithm \ref{alg:addmod}. Figure \ref{fig:results_add} show that AddRefine is extremely sample efficient, and can construct a new specification to aid the current edge with an extremely high success probability. We also show the distribution of states that make up the new specification $\phi_{goal}^r$.
To verify Algorithm \ref{alg:graph_refinement}, we designed a specification $\phi_{mid};\phi_{goal}$, as depicted in Figure \ref{fig:results_reach_past}. The learning curves, also shown in Figure \ref{fig:results_reach_past}, indicate that the proposed refinement significantly enhances the reach probability compared to the original specification. Additionally, Figure \ref{fig:results_reach_past} illustrates the distribution of states in the MID region from which the GOAL region can be reached, which informs the refinement process. Figure \ref{fig:pandagym} shows how sequential refinement can be applied on higher dimensional state spaces. Different refinements produce varying results, as shown in Figure \ref{fig:pandagym}.

\begin{figure}[!h]%
    \centering
    \subfloat{{\includegraphics[width=0.3\linewidth]{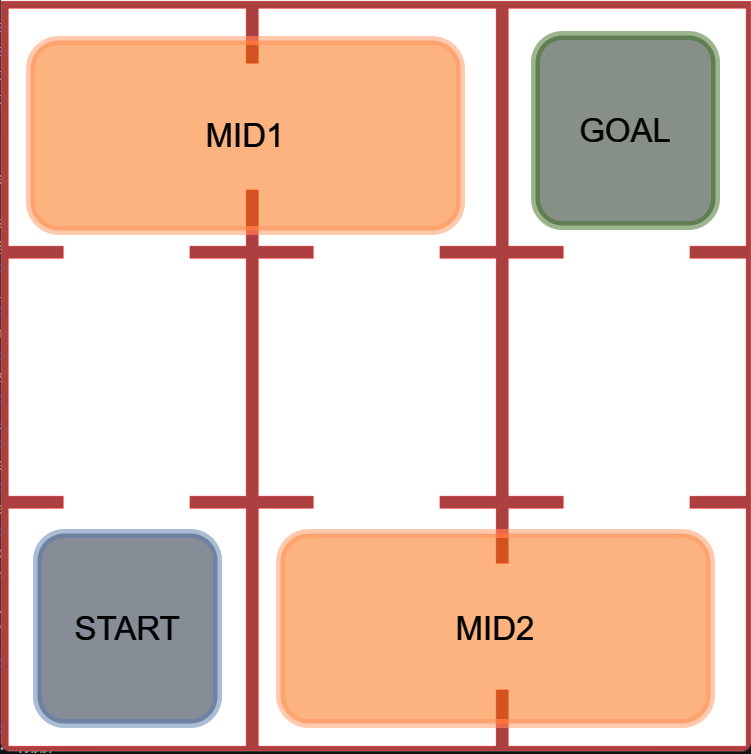} }}%
    \subfloat{{\includegraphics[width=0.4\linewidth]{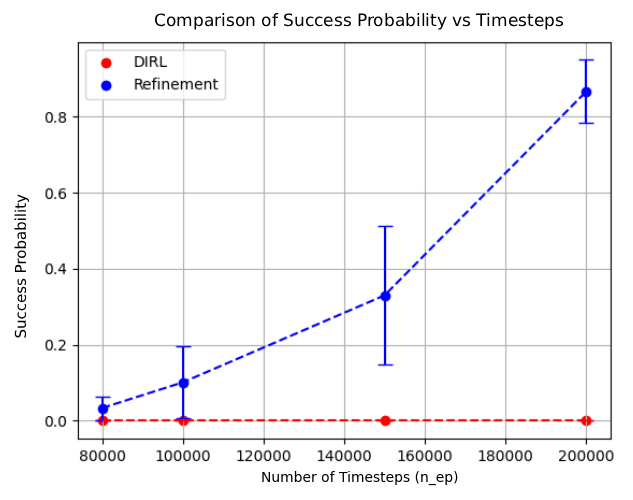} }}%
    
    \caption{Disjunctive Specification Refinement in 9-Rooms. (a) The environment with regions relevant to the specification \(\phi := \phi_{MID1}; \phi_{GOAL} \text{ or } \phi_{MID2}; \phi_{GOAL}\). (b) Learning curves showing that incorporating an additional specification \(\phi_{MID1}; \phi_{MID2}\) is essential to achieve the desired success probability. }
    
    \label{fig:results_or}%
\end{figure}
    
    

\paragraph{Experiment~3: Disjunctive refinement.} To validate Algorithm \ref{alg:dis_spec_refinement}, we constructed a 9 Rooms environment with a specification featuring two distinct paths to the goal. Figure \ref{fig:results_or} illustrates the environment and the regions relevant to the specification $\phi := \phi_{MID1}; \phi_{GOAL} \textbf{ or } \phi_{MID2}; \phi_{GOAL}$. The learning curves for \textbf{OrRefine}, shown in Figure \ref{fig:results_or}, demonstrate that incorporating an additional specification $\phi_{MID1}; \phi_{MID2}$ is essential to achieve good success probability. In contrast, Algorithm \ref{alg:seqmod} and Algorithm \ref{alg:graph_refinement} fail to perform adequately due to the subspecification having zero reach probability, preventing effective local refinement. This validation underscores the necessity of \textbf{OrRefine} in scenarios where sequential modifications alone are insufficient.



All Experiments have been performed using i7-8750H with 32GB RAM and no GPU. Trajectories were collected after training the policy for $n$ timesteps and 5 different seeds.

Hyperparameters for learning algorithms:

\begin{enumerate}
    \item Learning Rate: 0.0003
    \item n steps: 2048
    \item Batch size: 64
    \item Epochs: 10
    \item $\gamma$: 0.99
\end{enumerate}

\subsection{Limitations}

AutoSpec requires finite witnesses to specification satisfaction and hence can only work on finite trajectories. This means that we must consider only finitary fragments of LTL, like SpectRL. While Autospec is sound, i.e if a refinement is found satisfactory, trajectories for the refinement will also satisfy the original specification (Theorem 3.1), it is not complete, i.e. it might fail to find a candidate refinement even if such a refinement exists, especially if the specification satisfiability is extremely low. It is also not guaranteed that the candidate refinement is an 'optimal' refinement, in terms of the tightest bounds possible on the refined predicates.  

\subsection{Societal Impacts}

We wish to improve the performance of Reinforcement Learning algorithms and attempt to improve under-specified human specifications, which have an impact on various applications that aim to deploy RL agents with multi-objective tasks. The applications extend to robotics, path-finding tasks and any tasks that involve manual specifications which could be incorrect. This may have both positive or negative societal impacts depending on the use case of such RL deployments, positive impacts include applications to manufacturing, healthcare, and in-home robotic assistants; while negative impacts would be most consequential in military or surveillance infrastructure. These issues are shared across most work on reinforcement learning algorithms.

\end{document}